\documentclass[11pt]{article}
\usepackage{tcolorbox}
\usepackage{xcolor}
\usepackage[preprint]{acl}

\usepackage{times}
\usepackage{latexsym}
\usepackage{amsmath} 
\usepackage{amssymb} 
\usepackage[T1]{fontenc}

\usepackage[utf8]{inputenc}

\usepackage{microtype}

\usepackage{inconsolata}

\usepackage{graphicx}
\usepackage{placeins}

\usepackage{multirow}

\title{HELEA: Hard-Negative Benchmark and LLM-based Reranking for Robust Entity Alignment}

\author{Yoonjin Jang \\
  SungKyunKwan University \\
  Republic of Korea \\
  \texttt{yoonjinjang98@gmail.com} \\\And
  Junwoo Kim \\
  SungKyunKwan University \\
  Republic of Korea \\
  \texttt{junwookim618@gmail.com} \\\And
  Youngjoong Ko\thanks{Corresponding author} \\
  SungKyunKwan University \\
  Republic of Korea \\
  \texttt{yjko@skku.edu} \\}

\begin{document}
\maketitle

\begin{abstract}
    Entity Alignment (EA) is essential for knowledge graph (KG) fusion, but existing benchmarks often allow models to exploit name overlap rather than relational structure. This makes it difficult to evaluate whether models can reject same-name entities that refer to different real-world objects.
    Our primary contribution is a \textit{same-name hard-negative augmentation strategy} that simultaneously yields quality-controlled evaluation benchmarks (\textbf{DW-HN29K}, \textbf{DY-HN27K}) and augmented training corpora (\textbf{DW-Train}, \textbf{DY-Train}), by mining same-name but distinct entity pairs from KG name-collision groups.
    We further introduce \textbf{HELEA}, a two-stage framework integrating (i) entity encoder retrieval trained on hard-negative-augmented training corpora with 1-hop KG context, and (ii) LLM-based reranking without additional training.
    Experiments show that name-dependent baselines collapse to near-random performance on our hard-negative benchmarks, while HELEA achieves F1 0.967 on DW-HN29K while maintaining Hit@1 0.993 on standard DW-15K.\footnote{Our benchmark datasets and code are available at \url{https://github.com/Wnsdnl/HELEA}}
\end{abstract}

\section{Introduction}

Knowledge Graph (KG) fusion integrates multiple KGs into a unified resource, with Entity Alignment (EA) serving as its core mechanism. This process determines whether two entities from different KGs refer to the same real-world object~\cite{knowledge_graphs, hlmea_2025}.
Each entity in a KG has a \textit{name}---its human-readable label, such as a Wikipedia title---and a relational neighborhood that describes its surrounding context.
Because different KGs may represent the same object with different neighborhood structures and relational contexts, EA models must identify true alignments despite such structural discrepancies.

Most standard EA benchmarks were designed to evaluate whether models can recover positive alignments across different KGs.
Benchmarks such as \textbf{DW-15K} and \textbf{DY-15K}\footnote{Both the V1 density variant throughout.}~\cite{OpenEA} have played an important role in this setting.
They support the evaluation of cross-KG alignment between DBpedia--Wikidata and DBpedia--YAGO, and they are well suited to measuring whether models can match structurally related entities across heterogeneous KGs.
In this earlier setting, randomly sampled negatives were sufficient for evaluating the structural matching difficulty of EA.

However, EA evaluation now faces a different challenge as recent models increasingly exploit textual and structural signals.
They have become better at comparing entity descriptions and KG contexts, so benchmarks should test the subgraphs and structural contexts around entities instead of focusing mainly on their names.
Therefore, a benchmark should test the subgraphs and structural contexts around entities, instead of focusing mainly on their names.
However, current evaluations are hindered by a major shortcut: \textit{same-name entity pairs}.
In standard benchmarks, most positive pairs share identical names across KGs, while randomly sampled negative pairs rarely do.
For example, 81.71\% of positive pairs in \textbf{DW-15K} and 99.92\% in \textbf{DY-15K} share identical names.
As a result, a model can achieve high performance by relying on entity name similarity, without fully using relational structure.

This issue is especially important in KG fusion because the goal is not only to connect equivalent entities but also to keep distinct entities separate.
Entities may share the same or similar names while referring to different real-world objects.
For example, \texttt{Romeo\_and\_Juliet} may refer to a play, a film, or a ballet.
A model that relies mainly on name similarity may incorrectly merge such entities, even though their relational neighborhoods indicate different identities.
Same-name hard negatives directly represent this risk because they look similar at the name level, but they should not be aligned.
Standard benchmarks do not include such cases as a core evaluation setting.
Thus, they provide limited support for evaluating whether models can distinguish name-similar but semantically different entities.

\begin{figure*}[t]
    \centering
    \includegraphics[width=\textwidth]{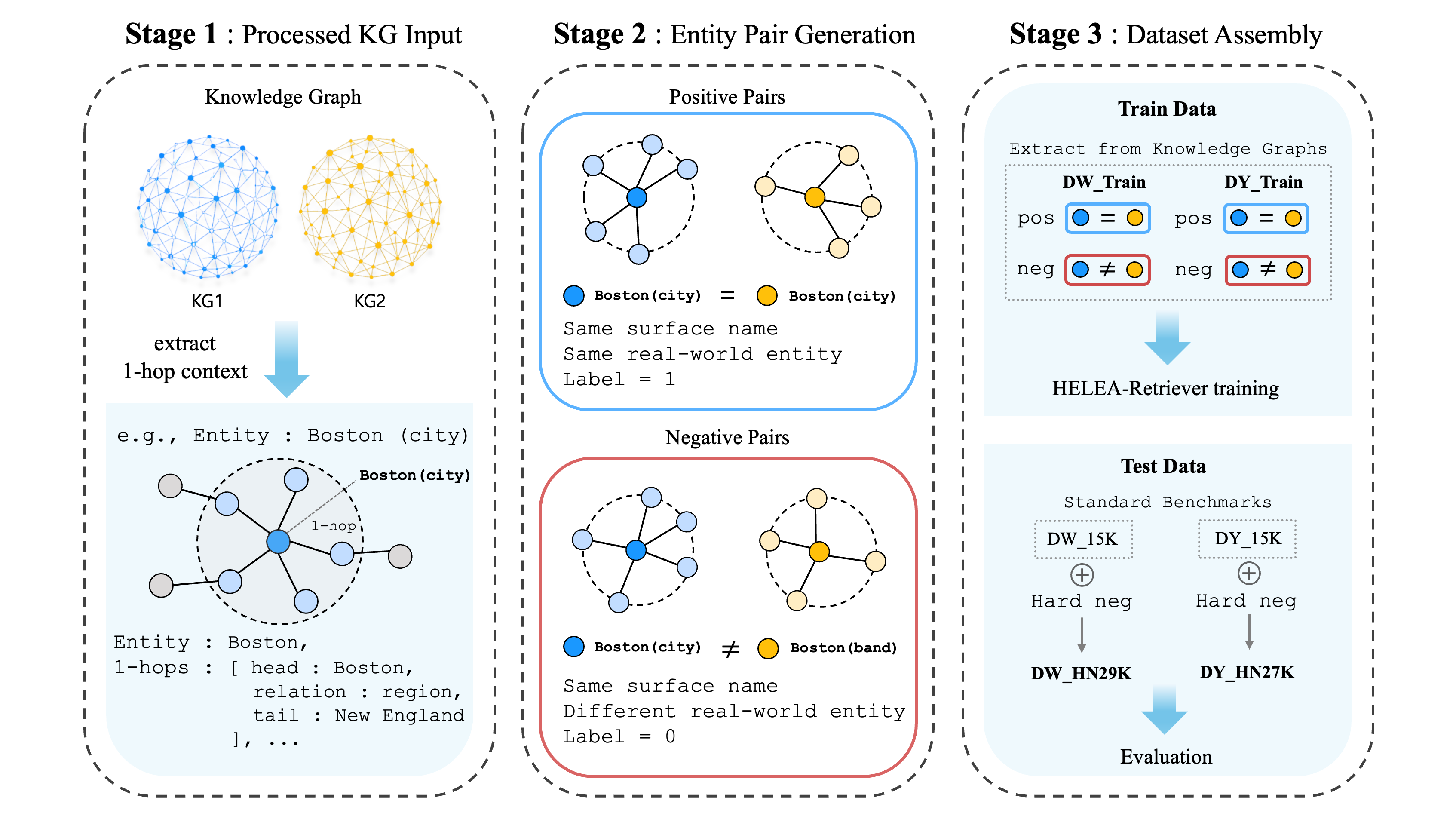}
    \caption{Overview of the same-name hard-negative pipeline. Same-name entity groups are mined from raw KG dumps, filtered through quality-control steps, and assembled into two outputs: augmented training corpora (DW-Train, DY-Train) used to train HELEA's entity encoder, and quality-controlled evaluation benchmarks (DW-HN29K, DY-HN27K).}
    \label{fig:benchmark}
\end{figure*}

To address this gap, we extend the standard \textbf{DW-15K} and \textbf{DY-15K} benchmarks with same-name hard negatives.
By keeping the original KG pairs and positive alignments, we directly measured how the evaluation changes when hard negatives are added.

This extension shifts the evaluation from simple positive alignment recovery to a more realistic KG fusion decision. A model must still identify entities that refer to the same real-world object, but it must also reject entities that share the same or similar names while referring to different objects. To succeed in this harder setting where entity names are no longer a reliable discriminative signal, models must compare 1-hop relational neighborhoods and decide whether the entity pair should actually be merged or not. To this end, we construct two quality-controlled hard-negative benchmarks, \textbf{DW-HN29K} and \textbf{DY-HN27K}.

We also propose \textbf{HELEA}, a two-stage EA framework for hard-negative-aware entity alignment.
HELEA first uses an entity encoder to retrieve candidate entities from serialized 1-hop KG contexts.
It then uses an LLM-based reranker to make a more informed decision among the retrieved candidates.

Our contributions are:
\begin{itemize}
  \item We identify and quantify the \textbf{name-bias shortcut} in standard EA benchmarks, showing that entity name overlap can allow simple matching strategies to perform strongly without using relational structure.
  \item We propose a \textit{same-name hard-negative augmentation strategy} that yields augmented training corpora (\textbf{DW-Train}, \textbf{DY-Train}) and quality-controlled evaluation benchmarks (\textbf{DW-HN29K}, \textbf{DY-HN27K}).
  \item We propose \textbf{HELEA}, a two-stage EA framework that combines entity encoder retrieval with LLM-based reranking. HELEA achieves F1 0.967 on \textbf{DW-HN29K} while maintaining Hit@1 0.993 on the standard \textbf{DW-15K}.
\end{itemize}

\section{Related Work}

\subsection{Benchmark Datasets for Entity Alignment}

To evaluate EA, various benchmark datasets have been proposed. Early benchmarks such as \textbf{WK3L}~\cite{MTransE_WK3L} and \textbf{DBP15K}~\cite{JAPE_DBP15K_DBP100K} focused on cross-lingual settings; \textbf{DWY100K}~\cite{BootEA_DWY100K} extended this to large-scale monolingual cross-KG benchmarks including DBpedia--Wikidata and DBpedia--YAGO~\cite{YAGO}. \textbf{SRPRS}~\cite{SRPRS} targeted more realistic degree distributions, and the \textbf{OpenEA} suite~\cite{OpenEA} systematized evaluation with cross-lingual (\textbf{EN-FR}, \textbf{EN-DE}) and cross-KG (\textbf{DW}, \textbf{DY}) datasets at 15K/100K scales with V1/V2 density variants. \textbf{DBP2.0}~\cite{DBP2.0} introduced dangling entities, and \textbf{ICEWS-WIKI}/\textbf{ICEWS-YAGO}~\cite{ICEWS} targeted highly heterogeneous KGs.

Despite these advances, none of the existing benchmarks include same-name but distinct entities as negatives, leaving the name-bias shortcut identified in Section~\ref{sec:name_bias} entirely unmeasured.

\subsection{Entity Alignment Methods}
Existing EA methods can be broadly divided into representation learning-based, GNN-based, and LLM-enhanced approaches.

\paragraph{Representation Learning-based Methods}
Representation learning-based methods map entities into a shared vector space. 
Specifically, translation-based mechanisms are utilized by \textbf{MTransE}~\cite{TransE, MTransE_WK3L}, while iterative bootstrapping is adopted by \textbf{BootEA}~\cite{BootEA_DWY100K} and \textbf{TransFlood}~\cite{TransFlood}. 
Additionally, semi- or self-supervised learning is employed in \textbf{MixTEA}~\cite{Mixtea} and \textbf{SelfKG}~\cite{SelfKG}, and side-information integration is leveraged by \textbf{BERT-INT}~\cite{Bertint}, \citet{Exploring}, and \textbf{UniEA}~\cite{UniEA}.

\paragraph{GNN-based Methods}
GNN-based methods capture structural and neighborhood information through graph neural networks. 
Specifically, \textbf{RDGCN}~\cite{RDGCN} introduces relation-aware dual-graph convolutions, while \textbf{RREA}~\cite{RREA} utilizes relational reflection transformations.

\paragraph{LLM-enhanced Methods}
Recently, LLM-enhanced methods have emerged to leverage the advanced reasoning capabilities of large language models. \textbf{ChatEA}~\cite{ChatEA} translates KG structures into natural language prompts for LLM inference, while \textbf{HLMEA}~\cite{hlmea_2025} frames EA as a multi-stage candidate filtering and selection task.

However, because these methods are evaluated on standard benchmarks lacking same-name hard negatives, their high performance heavily relies on name matching rather than a genuine understanding of relational structures.

\begin{table*}[!t]
    \centering
    \renewcommand{\arraystretch}{1.2}
    \resizebox{0.9\textwidth}{!}{%
    \begin{tabular}{l|cc|cc|cc|cc}
    \hline
    & \multicolumn{2}{c|}{\textbf{DW-15K}} & \multicolumn{2}{c|}{\textbf{DW-HN29K}} & \multicolumn{2}{c|}{\textbf{DY-15K}} & \multicolumn{2}{c}{\textbf{DY-HN27K}} \\
    \textbf{Baseline} & \textbf{Hit@1} & \textbf{MRR} & \textbf{Acc.} & \textbf{F1} & \textbf{Hit@1} & \textbf{MRR} & \textbf{Acc.} & \textbf{F1} \\ \hline
    String-matching             & 0.825 & 0.826 & 0.411 & 0.579 & 0.999 & 0.999 & 0.457 & 0.628 \\
    Substring-matching & 0.844 & 0.866 & 0.445 & 0.615 & 1.000 & 1.000 & 0.457 & 0.628 \\
    Name-only Bi-encoder               & 0.899 & 0.918 & 0.495 & 0.662 & 0.993 & 0.996 & 0.457 & 0.628 \\ \hline
    \end{tabular}%
    }
    \caption{Performance of name-dependent baselines across all benchmarks. The high scores on standard benchmarks contrasted with the performance collapse on our HN benchmarks clearly demonstrate the severity of name bias.}
    \label{tab:name_bias_perf}
\end{table*}

\section{Hard-Negative Augmentation Strategy and Benchmark Construction}
\label{sec:benchmark}

\subsection{Name Bias in Existing Entity Alignment Benchmarks}
\label{sec:name_bias}

We use the standard cross-KG EA benchmarks, \textbf{DW-15K} (DBpedia~\cite{mendes-etal-2012-dbpedia}--Wikidata~\cite{wikidata}) and \textbf{DY-15K} (DBpedia--YAGO~\cite{YAGO}), as the base benchmarks for our extension.
Instead of constructing new KGs from scratch, we augment the original datasets by introducing hard negative pairs while keeping the positive alignments.
This design allows us to isolate the effect of same-name hard negatives while preserving the standard positive-alignment setting used in prior EA studies.

As shown in Table~\ref{tab:name_overlap} (Appendix~\ref{app:stats}), 81.71\% of positive pairs in \textbf{DW-15K} share an identical name between the two KGs (e.g., \texttt{Jim\_Jones} from DBpedia $\leftrightarrow$ \texttt{Jim\_Jones} from Wikidata). This pattern is even more pronounced in \textbf{DY-15K}, which exhibits a \textbf{99.92\%} name overlap. Because randomly sampled negatives rarely share the name, simple name-lookup baselines achieve misleadingly high performance on standard benchmarks (Section~\ref{sec:validation}).

\subsection{Same-Name Hard-Negative Augmentation Strategy}
\label{sec:hard_neg}
We propose a \textit{same-name hard-negative augmentation strategy} for EA, built on the observation that knowledge graphs naturally contain name-collision groups---sets of entities sharing an identical name that refer to different real-world objects (e.g., \texttt{Romeo\_and\_Juliet} as a play, a film, or a ballet).
These groups span both cross-KG pairs (e.g., a DBpedia entity and a same-name Wikidata entity of a different type) and within-KG pairs (e.g., two Wikidata entities sharing a name).
We extract these groups by normalizing entity labels and grouping entries with the same name, as detailed in Table~\ref{tab:benchmark_stats}.
Same-name pairs referring to the same object become positive samples; same-name pairs referring to different objects become hard-negative samples.

\subsection{Construction and Augmentation Pipeline}
\label{sec:pipeline}
As illustrated in Figure~\ref{fig:benchmark}, we instantiate the strategy above via a three-stage pipeline that simultaneously produces augmented training corpora and quality-controlled evaluation benchmarks from raw KG dumps.

\paragraph{Stage 1: Processed KG Input.} We extract 1-hop context per entity from KG dumps, retaining English-tagged triples. Entities whose names are bare QIDs (e.g., \texttt{Q12345}) and triples containing unresolved QIDs are discarded; entities left with an empty neighborhood after cleaning are excluded. Entity names are canonicalized by stripping parenthetical suffixes (e.g., \texttt{Boston (city)} $\rightarrow$ \texttt{Boston}) and applying Unicode normalization, ensuring accurate name-collision group identification.

\paragraph{Stage 2: Entity Pair Generation.} The augmentation strategy is applied in two modes. For \textit{training}, same-name pairs are drawn from the full KG dumps: pairs referring to the same real-world object serve as positives, while pairs referring to different objects constitute the injected hard negatives. For \textit{evaluation}, positives are taken from the standard seed alignments (\textbf{DW-15K}, \textbf{DY-15K}) and hard negatives from the name-collision groups of those specific seed entities.

\paragraph{Stage 3: Dataset Assembly.} For \textit{Augmented Training Corpora}, the mined pairs yield \textbf{DW-Train} (5,568,150 samples) and \textbf{DY-Train} (1,996,008 samples). For \textit{Quality-Controlled Benchmarks}, these are assembled into \textbf{DW-HN29K} and \textbf{DY-HN27K}.
To prevent data leakage, we filter all training samples that appear in the benchmark evaluation sets, ensuring no test entity is seen during training.

\subsection{Benchmark Statistics}
\label{sec:stats}
Our new DW benchmark, \textbf{DW-HN29K}, contains 29,718 samples (14,718 positive alignments and 15,000 hard negatives) with 1-hop context. It is derived from an initial set of 30,000 samples after applying the Stage~1 cleaning criteria: 282 pairs were removed because at least one entity in each pair had an empty 1-hop neighborhood or only unresolved QID references after cleaning. The DY benchmark, \textbf{DY-HN27K}, contains 27,635 samples (12,635 positives and 15,000 hard negatives). Full statistics are in Table~\ref{tab:benchmark_stats} (Appendix~\ref{app:stats}).
\begin{figure*}[t]
    \centering
    \includegraphics[width=\textwidth]{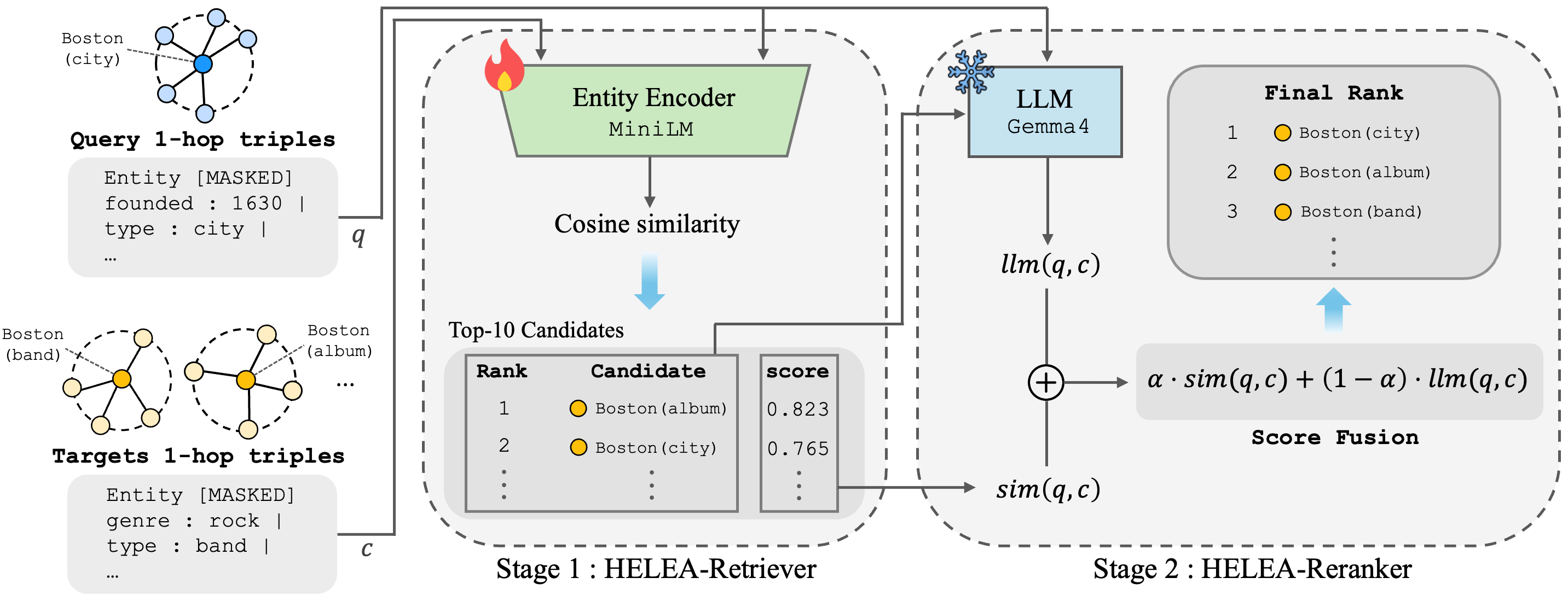}
    \caption{Overview of the HELEA pipeline.
      \textbf{Stage~1 (HELEA-Retriever):} Entities are serialized as 1-hop KG triples and encoded by a shared entity encoder; top-$K$ candidates are retrieved by cosine similarity.
      \textbf{Stage~2 (HELEA-Reranker):} The HELEA-Reranker reranks the candidates; entity encoder and reranker scores are linearly fused for the final decision.}
    \label{fig:pipeline}
\end{figure*}

\subsection{Validating the Name Bias Hypothesis}
\label{sec:validation}
To confirm that our benchmarks eliminate the name-bias shortcut, we evaluate three name-dependent baselines across all datasets as shown in Table~\ref{tab:name_bias_perf}.

\paragraph{String-matching baseline.} This baseline predicts a match if names are identical. On \textbf{DW-15K}, it covers 82.5\% of queries with almost no false positives due to random negative sampling, yielding misleadingly high performance. On \textbf{DW-HN29K}, however, because most of the hard negatives share the same name as their positive counterpart, the baseline generates numerous false positives and collapses to an accuracy of 0.411 and an F1 score of 0.579---performing barely above random chance. DY shows the same pattern (Hit@1 0.999 on \textbf{DY-15K}, F1 0.628 on \textbf{DY-HN27K}).

\paragraph{Substring matching baseline.} This baseline extends string-matching with a substring fallback. On \textbf{DW-15K}, this marginally improves over pure string-matching (Hit@1 0.844 vs.\ 0.825), but the collapse on \textbf{DW-HN29K} is similar---accuracy 0.445, F1 0.615. DY results are consistent (Hit@1 1.000 on \textbf{DY-15K}, F1 0.628 on \textbf{DY-HN27K}).

\paragraph{Name-only Bi-encoder baseline.} We evaluate a neural name-matching model (all-MiniLM-L6-v2~\citealt{minilm}) using only entity names as input. Binary predictions on hard-negative benchmarks are made by thresholding the cosine similarity at the value that maximizes F1 on the validation set. On \textbf{DW-15K}, it achieves Hit@1 = 0.899 and MRR = 0.918, outperforming several structure-aware methods. On \textbf{DW-HN29K}, however, it assigns nearly identical cosine similarities to hard negatives and true positives; since no score cutoff can separate them, it effectively predicts all pairs as positive (F1 0.662).

\section{Methodology}
\label{sec:model}

We propose \textbf{HELEA} (\textbf{H}ard-negative \textbf{E}nhanced entity encoder with \textbf{L}LM-based reranking for \textbf{E}ntity \textbf{A}lignment), a two-stage framework consisting of (i) \textbf{HELEA-Retriever}: an entity encoder trained on hard-negative-augmented training corpora (Section~\ref{sec:dpr}), and (ii) \textbf{HELEA-Reranker}: an LLM-based reranker applied without further training (Section~\ref{sec:reranker}).
An overview of the full pipeline is shown in Figure~\ref{fig:pipeline}.

\subsection{HELEA-Retriever}
\label{sec:dpr}

We employ an entity encoder built on
all-MiniLM-L6-v2~\cite{minilm, reimers-gurevych-2019-sentence}, a 6-layer BERT variant with a 384-dimensional hidden state. Each entity $e$ is represented as a serialized 1-hop KG context string $\text{rep}(e)$, formed by prepending the entity name to its deduplicated 1-hop triple sequence (e.g., \texttt{Boston | birthPlace: John Adams | ...}).
\begin{equation}
  \mathbf{e} = \frac{f_\theta(\text{rep}(e))}{\|f_\theta(\text{rep}(e))\|_2}
  \label{eq:encoder}
\end{equation}
where $f_\theta$ extracts the CLS token representation from the final hidden layer. The structural similarity between two entities $a$ and $b$ is computed as the cosine similarity $s_{ab} = \mathbf{e}_a^\top \mathbf{e}_b$.

We train the retriever using a bidirectional InfoNCE loss~\cite{DBLP:journals/corr/abs-1807-03748}:
\begin{align}
  \mathcal{L} &= \frac{1}{2}\bigl(\mathcal{L}_{A \to B} + \mathcal{L}_{B \to A}\bigr) \label{eq:bidi_nce} \\
  \mathcal{L}_{A \to B} &= -\frac{1}{|P|}\sum_{i \in P} \log \frac{\exp(\tau \cdot s_{ii})}{\sum_{j=1}^{N} \exp(\tau \cdot s_{ij})} \label{eq:nce_standard}
\end{align}
where $P$ is the set of positive-pair indices within a batch of size $N$, and $s_{ij} = (\mathbf{e}_i^A)^\top \mathbf{e}_j^B$. Here, $\tau = \exp(\log\text{-scale})$ where $\log\text{-scale}$ is a learnable parameter.
Non-positive pairs within each batch (where label $= 0$) are excluded from $P$ and naturally serve as in-batch hard negatives in the denominator. During inference, all target candidates are encoded and indexed with FAISS~\cite{faiss}. The top-$K$ entities are retrieved by cosine similarity against the query embedding.

\begin{table*}[t]
  \centering
  \renewcommand{\arraystretch}{1.2}
  \resizebox{\textwidth}{!}{%
  \begin{tabular}{l|cc|cccc|cc|cccc}
  \hline
  & \multicolumn{2}{c|}{\textbf{DW-15K}} & \multicolumn{4}{c|}{\textbf{DW-HN29K}} & \multicolumn{2}{c|}{\textbf{DY-15K}} & \multicolumn{4}{c}{\textbf{DY-HN27K}} \\
  \textbf{Method} & \textbf{Hit@1} & \textbf{MRR} & \textbf{Acc.} & \textbf{Prec.} & \textbf{Recall} & \textbf{F1} & \textbf{Hit@1} & \textbf{MRR} & \textbf{Acc.} & \textbf{Prec.} & \textbf{Recall} & \textbf{F1} \\ \hline
  \multicolumn{13}{l}{\textit{Name-dependent baselines}} \\ \hline
  String-matching      & 0.825 & 0.826 & 0.411 & 0.448 & 0.820 & 0.579 & 0.999 & 0.999 & 0.457 & 0.457 & 1.000$^\dagger$ & 0.628 \\
  Name-only Bi-encoder & 0.899 & 0.918 & 0.495 & 0.495 & 1.000$^\dagger$ & 0.662 & 0.993 & 0.996 & 0.457 & 0.457 & 1.000$^\dagger$ & 0.628 \\ \hline
  \multicolumn{13}{l}{\textit{Reproduced baselines}} \\ \hline
  BERT-INT$^\|$~\cite{Bertint}        & \textbf{0.994} & \textbf{0.994} & 0.626 & 0.557 & 0.901 & 0.688 & \textbf{0.999} & \textbf{0.999} & 0.736 & 0.703 & 0.732 & 0.717 \\
  SelfKG$^\S$~\cite{SelfKG}          & 0.920 & 0.931 & 0.727 & 0.692 & 0.812 & 0.747 & \underline{0.998} & \underline{0.999} & 0.689 & 0.639 & 0.797 & 0.709 \\
  ChatEA~\cite{ChatEA}           & 0.955 & 0.961 & \underline{0.956} & \underline{0.856} & \underline{0.982} & \underline{0.915} & 0.980 & 0.983 & \underline{0.927} & \underline{0.918} & \underline{0.923} & \underline{0.920} \\
  UniEA$^\#$~\cite{UniEA}            & 0.901 & 0.919 & 0.568 & 0.537 & 0.933 & 0.681 & 0.760 & 0.814 & 0.574 & 0.521 & 0.864 & 0.650 \\
  HLMEA$^\ddagger$~\cite{hlmea_2025} & 0.842 & 0.877 & 0.852 & 0.819 & 0.900 & 0.858 & 0.989 & 0.993 & 0.654 & 0.578 & 0.903 & 0.705 \\ \hline
  \multicolumn{13}{l}{\textit{Ours}} \\ \hline
  HELEA-Retriever & 0.987 & 0.991 & 0.847 & 0.815 & 0.895 & 0.853 & 0.992 & 0.995 & 0.723 & 0.670 & 0.776 & 0.719 \\
  \textbf{HELEA}  & \underline{0.993} & \underline{0.994} & \textbf{0.968} & \textbf{0.973} & \textbf{0.961} & \textbf{0.967} & 0.992 & 0.995 & \textbf{0.937} & \textbf{0.925} & \textbf{0.938} & \textbf{0.931} \\ \hline
  \end{tabular}%
  }
  \caption{Results on DW-15K and DY-15K (retrieval: Hit@1, MRR) and DW-HN29K and DY-HN27K (binary classification: Acc., Prec., Recall, F1).
    $^\dagger$Recall\,=\,1.0: degenerate always-positive prediction.
    $^\|$BERT-INT accesses full entity descriptions beyond the 1-hop triples provided to all other methods. HN benchmark results additionally fine-tune on DW-Train/DY-Train hard negatives.
    $^\S$For HN benchmarks, SelfKG uses frozen embeddings with a separate LR classifier trained on DW-Train/DY-Train.
    $^\#$UniEA is retrained transductively on each benchmark's 2/1/7 split; as a GNN it cannot leverage external corpora such as DW-Train or DY-Train.
    $^\ddagger$HLMEA is self-tuned on each HN benchmark following the HLMEA framework.
    \textbf{Bold}/\underline{underline}: best/second-best by MRR (15K benchmarks) and F1 (HN benchmarks).}
  \label{tab:exp1}
\end{table*}

\subsection{HELEA-Reranker and Score Fusion}
\label{sec:reranker}

For each query, the top-$K$ retrieved candidates are submitted in a single \textit{listwise} prompt alongside their names and 1-hop KG triples (full template in Appendix~\ref{app:prompt}). The LLM outputs a ranked list of \texttt{index:score} pairs, where each score is a confidence value clamped to $[0,1]$. We use this resulting value as the LLM score component $\text{llm}(q, c)$.
This listwise design incurs exactly one LLM call per query regardless of candidate pool size, making HELEA substantially more inference-efficient than pairwise or multi-round approaches such as ChatEA, which require $O(K)$ sequential LLM interactions per query.

The final entity alignment score is formulated as a linear fusion of the cosine similarity $\text{sim}(q, c) \in [-1, 1]$ and the LLM reranking score:
\begin{equation}
  \text{score}(q, c) = \alpha \cdot \text{sim}(q, c) + (1 - \alpha) \cdot \text{llm}(q, c)
  \label{eq:fusion}
\end{equation}
where the weighting hyperparameter $\alpha$ is task-dependent and optimized via ablation studies (Section~\ref{sec:ablation_alpha}). These fused scores determine the final rankings for Hit@$K$ and MRR evaluations on the standard \textbf{DW-15K} and \textbf{DY-15K} datasets. For the hard-negative benchmarks (\textbf{DW-HN29K} and \textbf{DY-HN27K}), the fused score for the target evaluation pair is extracted from the listwise output and thresholded for binary prediction, with threshold selection described in Section~\ref{sec:experimental_setup}.

\section{Experiments}
\label{sec:experiments}

\subsection{Experimental Setup}
\label{sec:experimental_setup}

\paragraph{Datasets.}
We evaluate on two KG pairs, each with a positive-only and a hard-negative benchmark,
all annotated with 1-hop KG triples.
For DBpedia--Wikidata (DW): \textbf{DW-15K} (15,000 positive pairs from~\cite{OpenEA})
and \textbf{DW-HN29K} (Section~\ref{sec:benchmark}).
For DBpedia--YAGO (DY): \textbf{DY-15K} (15,000 positive pairs from~\cite{OpenEA})
and \textbf{DY-HN27K} (12,635 positive + 15,000 same-name hard negatives).

\paragraph{Evaluation metrics.}
On DW-15K and DY-15K we report Hit@1 and MRR, ranking each entity$_a$ query against all entity$_b$ candidates in the 10,500-pair held-out test set (20/10/70 split).
On \textbf{DW-HN29K} and \textbf{DY-HN27K} we report \textbf{binary classification} (Accuracy, Precision, Recall, F1) and \textbf{retrieval with a hard-negative pool} (Hit@1, Table~\ref{tab:hn_retrieval}), where the candidate pool contains every row of the benchmark so each positive query faces at least one same-name distractor.
The decision threshold is swept over the held-out 10K validation set to maximize F1 and applied fixed to the evaluation set.

\paragraph{Baselines.}
We compare against \textbf{UniEA}~\cite{UniEA}, \textbf{BERT-INT}~\cite{Bertint},
\textbf{SelfKG}~\cite{SelfKG}, \textbf{ChatEA}~\cite{ChatEA}, and
\textbf{HLMEA}~\cite{hlmea_2025}.
We also include a \textbf{Name-only Bi-encoder} (names only, no triples) as a direct measure of name bias, and \textbf{HELEA-Retriever} to isolate entity encoder performance from reranking.

\paragraph{Implementation details.}
Full hyperparameters, validation split procedure, and LLM serving configuration are in Appendix~\ref{app:impl}.
Briefly, we fine-tune all-MiniLM-L6-v2~\cite{minilm, reimers-gurevych-2019-sentence} as an entity encoder and rerank its top-10 candidates with Gemma 4 31B Instruct~\cite{gemma4_2025} via vLLM~\cite{kwon2023efficient}.
The fusion weight is $\alpha = 0.75$ for retrieval benchmarks and $\alpha = 0.25$ for hard-negative benchmarks, as determined by ablation in Section~\ref{sec:ablation_alpha}; GPT-oss-120B~\cite{gpt_oss_120b} experiments confirming backbone generalizability appear in Appendix~\ref{app:gpt_oss}.

\begin{table}[t]
  \centering
  \renewcommand{\arraystretch}{1.2}
  \resizebox{0.85\columnwidth}{!}{%
  \begin{tabular}{lcc}
    \hline
    \textbf{Method} & \textbf{DW-HN29K} & \textbf{DY-HN27K} \\
                    & \textbf{Hit@1}    & \textbf{Hit@1}    \\ \hline
    BERT-INT        & 0.934             & 0.897             \\
    SelfKG          & 0.897             & 0.897             \\
    HLMEA           & 0.669             & 0.872             \\ \hline
    HELEA-Retriever & 0.949             & 0.889             \\
    \textbf{HELEA}  & \textbf{0.959}    & \textbf{0.923}    \\ \hline
  \end{tabular}
  }
  \caption{Hit@1 on DW-HN29K and DY-HN27K in the \textit{retrieval} setting.
  Every positive query faces at least one same-name distractor in the pool.
  ChatEA uses pairwise LLM scoring and cannot rank a full candidate pool without $O(N)$ LLM calls per query; it is excluded from this table.}
  \label{tab:hn_retrieval}
\end{table}

\subsection{Results on DW-15K and DY-15K}
\label{sec:exp2}

Table~\ref{tab:exp1} reports results on all four benchmarks.
On standard benchmarks, \textbf{HELEA} is competitive across both KG pairs, reaching Hit@1 0.993 on \textbf{DW-15K}---within 0.1\%p of \textbf{BERT-INT}, which supplements 1-hop triples with richer entity descriptions.
\textbf{DY-15K} is effectively a name-matching test (99.9\% name overlap), where string-matching itself achieves Hit@1 0.999; the small gap between \textbf{HELEA} (Hit@1 0.992) and \textbf{BERT-INT} (0.999) reflects YAGO's sparse triple context (avg.\ 8.79 vs.\ 28.53 triples for Wikidata), making description-based methods relatively more advantageous.

\subsection{Results on DW-HN29K and DY-HN27K}
\label{sec:exp3}

On \textbf{DW-HN29K}, name-dependent models effectively predict all pairs as positive---both the Name-only Bi-encoder and unmodified \textbf{BERT-INT} plateau near the always-positive F1 ceiling of 0.66 (Table~\ref{tab:exp1}).
Hard-negative fine-tuning moves \textbf{BERT-INT} only marginally: because name similarity dominates its interaction-based scoring, same-name entities receive nearly identical representations regardless of their relational context, and the improvement is less than 0.03 F1.
Structure-aware methods recover substantially, with \textbf{HLMEA} and \textbf{HELEA-Retriever} both reaching F1~0.85.
\textbf{HELEA} adds a further \textbf{+11.4\%p} by replacing the entity encoder's name-influenced similarity scores with the \textbf{HELEA-Reranker}'s triple-by-triple semantic comparison, reaching F1 0.967.
Among reproduced baselines, \textbf{ChatEA} comes closest (F1 0.915) by feeding full entity triples and descriptions into the LLM through a multi-round process---an approach that naturally disambiguates same-name entities without explicit hard-negative training, though at substantially higher per-query inference cost.

The pattern is more pronounced on \textbf{DY-HN27K}, where every hard negative shares an exact name with its positive pair, leaving the \textbf{HELEA-Retriever} with no name signal to anchor ranking (F1 0.719).
The \textbf{HELEA-Reranker}'s listwise triple comparison is therefore mainly responsible for recovery, lifting the final score to F1 0.931---a \textbf{+21.2\%p} gain over \textbf{HELEA-Retriever}.

\paragraph{Retrieval under a hard-negative pool.}
Table~\ref{tab:hn_retrieval} evaluates baseline methods in a retrieval setting where the candidate pool contains every row of \textbf{DW-HN29K} and \textbf{DY-HN27K} (29,718 and 27,635 entries respectively), so that each positive query must be ranked above at least one same-name distractor.
On \textbf{DW-HN29K}, name-reliant methods suffer the most. \textbf{HLMEA} drops from 0.842 on \textbf{DW-15K} to 0.669 Hit@1, while \textbf{HELEA} achieves the highest Hit@1 of 0.959, surpassing \textbf{HELEA-Retriever} (0.949), \textbf{BERT-INT} (0.934), and \textbf{SelfKG} (0.897).

This stands in apparent contrast to \textbf{BERT-INT}'s near-chance binary accuracy (Table~\ref{tab:exp1}). Because it assigns nearly identical scores to positive and same-name negative pairs, relative ordering is often preserved and Hit@1 remains high. However, no global threshold can separate the two score distributions.

On \textbf{DY-HN27K}, where 100\% of hard negatives share an exact name with the positive, all baselines cluster between 0.872 and 0.897 Hit@1; \textbf{HELEA} rises to 0.923, a +3.4\%p gain over \textbf{HELEA-Retriever} (0.889), confirming that the \textbf{HELEA-Reranker} provides meaningful separation even under complete name overlap.

\begin{table*}[!t]
\centering
\renewcommand{\arraystretch}{1.2}
\resizebox{0.8\textwidth}{!}{%
\begin{tabular}{l|cc|ccc|cc|ccc}
\hline
& \multicolumn{2}{c|}{\textbf{DW-15K}} & \multicolumn{3}{c|}{\textbf{DW-HN29K (Val.)}} & \multicolumn{2}{c|}{\textbf{DY-15K}} & \multicolumn{3}{c}{\textbf{DY-HN27K (Val.)}} \\
\textbf{$\alpha$} & \textbf{Hit@1} & \textbf{MRR} & \textbf{Acc.} & \textbf{F1} & \textbf{Thr.} & \textbf{Hit@1} & \textbf{MRR} & \textbf{Acc.} & \textbf{F1} & \textbf{Thr.} \\ \hline
1.00 (Retriever) & 0.9961          & 0.9979          & 0.717          & 0.741          & 0.32 & 0.9987          & 0.9991          & 0.848          & 0.902          & 0.46 \\
\textbf{0.75}    & 0.9974          & 0.9987          & 0.774          & 0.782          & 0.48 & \textbf{0.9993} & \textbf{0.9994} & 0.896          & 0.932          & 0.56 \\
0.50             & 0.9987          & 0.9994          & 0.837          & 0.836          & 0.62 & 0.9993          & 0.9994          & 0.922          & 0.948          & 0.68 \\
\textbf{0.25}    & \textbf{0.9987} & \textbf{0.9994} & \textbf{0.888} & \textbf{0.881} & 0.78 & 0.9987          & 0.9991          & \textbf{0.924} & \textbf{0.949} & 0.78 \\
0.00 (Reranker)       & 0.9987          & 0.9994          & 0.867          & 0.867          & 0.86 & 0.9987          & 0.9991          & 0.921          & 0.947          & 0.82 \\ \hline
\end{tabular}%
}
\caption{Ablation on score fusion weight $\alpha$,
          evaluated on held-out validation sets: 1,497-pair fold-1 valid split (Hit@K) and 10K held-out validation set (HN Acc/F1).
          $\alpha{=}1.0$: HELEA-Retriever score only; $\alpha{=}0.0$: LLM score only.
          \textbf{Thr.} is the decision threshold on fused scores that maximizes F1 on the validation set, applied fixed to the evaluation set.}
\label{tab:exp5}
\end{table*}

\subsection{Ablation Studies}
\label{sec:ablation}

\paragraph{Score Fusion Weight $\alpha$.}
\label{sec:ablation_alpha}

Table~\ref{tab:exp5} sweeps the fusion weight $\alpha$ (Equation~\ref{eq:fusion}) on the held-out 10K validation sets (randomly excluded from the
DW-Train and DY-Train training corpora); HELEA-Retriever top-10 candidates are fixed throughout.

The central finding is that even a small \textbf{HELEA-Reranker} contribution is sufficient to override name-driven scoring: \textbf{HELEA-Retriever} alone is dominated by name similarity, but any nonzero LLM weight consistently improves hard-negative discrimination.

On standard retrieval benchmarks, the retriever already performs well (Hit@1 0.996/0.999 DW/DY), and both datasets reach near-ceiling by $\alpha{=}0.50$ and plateau thereafter.
The effect is more pronounced on HN benchmarks. The retriever scores same-name hard negatives nearly as high as true positives at $\alpha{=}1.0$, and any LLM weight replaces these name-driven scores with triple-based semantic judgments.

Both HN benchmarks improve steadily as $\alpha$ decreases; \textbf{DW-HN29K} reaches F1 0.881 at $\alpha{=}0.25$ (\textbf{+14.0\%p} over retriever-only), with full per-$\alpha$ trends in Table~\ref{tab:exp5}.
At $\alpha{=}0.00$, which means reranking with pure LLM score, \textbf{DW-HN29K} drops slightly to F1 0.867, confirming that score fusion outperforms the reranker alone.

We adopt $\alpha{=}0.75$ for retrieval and $\alpha{=}0.25$ for HN benchmarks: on retrieval benchmarks the retriever score is already discriminative (randomly sampled negatives rarely share a name), so the reranker contributes at the margin; on HN benchmarks the cosine gap between same-name negatives (${\sim}0.9999$) and true positives (${\sim}0.9608$) is too small to threshold reliably, and the reranker's triple-based confidence must dominate.
Backbone generalizability is further confirmed with GPT-oss-120B in Appendix~\ref{app:gpt_oss}.

\paragraph{Training Data: Hard Negatives vs.\ Scale.}
\label{sec:ablation_data}

Hard-negative signal, not data volume, drives the gain (Table~\ref{tab:ablation_data}).
At the same 1.1M sample count, switching from positive-only to \textbf{DW-Train} 20\% raises \textbf{DW-HN29K} F1 by \textbf{+2.8\%p}.
Scaling beyond 20\% yields no further improvement (Accuracy plateaus at 86.8--87.1\%), indicating that hard-negative diversity is well-covered by the first random 20\% of \textbf{DW-Train}; a competitive \textbf{HELEA-Retriever} can therefore be trained on $\sim$1M augmented samples---a 5$\times$ reduction with negligible accuracy cost.
Notably, \textbf{DW-15K} Hit@1 remains stable across all data fractions, confirming that hard-negative training does not compromise standard retrieval performance.
We release the full \textbf{DW-Train} to support future work, as larger models or different architectures may benefit from additional hard-negative diversity beyond this threshold.

\begin{table}[t]
\centering
\renewcommand{\arraystretch}{1.2}
\resizebox{0.95\columnwidth}{!}{%
\begin{tabular}{lrcc}
\hline
\textbf{Training data} & \textbf{Samples} & \textbf{Acc.} & \textbf{F1} \\
\hline
Positive-only (20\%) & 1.1M & 0.845 & 0.849 \\
\hline
DW-Train 20\% & 1.1M & 0.869 & 0.877 \\
DW-Train 40\% & 2.2M & 0.869 & 0.875 \\
DW-Train 60\% & 3.3M & 0.871 & 0.876 \\
DW-Train 80\% & 4.5M & 0.870 & 0.875 \\
DW-Train 100\% & 5.6M & 0.868 & 0.876 \\
\hline
\end{tabular}%
}
\caption{Ablation on training data composition and scale, evaluated on DW-HN29K.
The top row trains on positive pairs only (no hard negatives), at the same data volume as the 20\% condition.
Hard-negative augmentation does not degrade standard retrieval: HELEA-Retriever (DW-Train 100\%) achieves DW-15K Hit@1 0.987 (Table~\ref{tab:exp1}).}
\label{tab:ablation_data}
\end{table}

\section{Conclusion}
\label{sec:conclusion}

We addressed a fundamental gap in cross-KG entity alignment evaluation, where standard benchmarks can be largely solved by name lookup alone, masking whether models genuinely learn from relational structure.
To expose this gap, we constructed \textbf{DW-HN29K} and \textbf{DY-HN27K}, quality-controlled hard-negative benchmarks where 99\%+ of negatives share an exact name with their paired positive, causing all name-matching baselines to collapse.
To meet this challenge, we proposed \textbf{HELEA}, a two-stage framework combining entity encoder retrieval trained on \textbf{DW-Train} with LLM reranking, achieving F1 0.967 on \textbf{DW-HN29K} while maintaining Hit@1 0.993 on standard \textbf{DW-15K}.

\section*{Limitations}
\textbf{HELEA} depends on LLM-based reranking, and fewer than 0.4\% of LLM calls return malformed outputs and fall back to \textbf{HELEA-Retriever} ranking (0.01\% on DW, 0.30\% on DY).
This rate is low, but more robust output control could further reduce fallback cases.

Another limitation is the sparsity of KG context.
On \textbf{DY-15K}, YAGO contains only 8.79 triples per entity on average, which limits structural evidence and contributes to the performance gap against description-based methods such as \textbf{BERT-INT}, as discussed in Section~\ref{sec:exp2}.
Extending context from 1-hop to 2-hop neighborhoods may alleviate this issue for sparse KGs.

For \textbf{DY-HN27K}, the binary-classification threshold is tuned on a held-out split of \textbf{DY-Train}, which is more positive-heavy ($\sim$73\% positive) than \textbf{DY-HN27K} ($\sim$46\% positive).
This distribution gap may bias predictions toward the positive class and inflate Recall at the expense of Precision.
More balanced validation splits or benchmark-specific calibration could reduce this effect.

Finally, our same-name hard-negative mining is currently applied to monolingual cross-KG benchmarks.
Extending it to cross-lingual benchmarks such as DBP15K(EN-FR) is possible, although lower surface-form overlap may make name bias less acute.
Hard-negative-aware InfoNCE training and in-batch same-name sampling are also promising directions for closing the gap between HELEA-Retriever and HELEA-Reranker.
We release all data, code, and checkpoints to support further work on name-robust entity alignment.

\section*{Ethical Considerations}

This work uses public KG artifacts under their respective licenses and terms of use: Wikidata structured data under CC0, DBpedia dumps under Creative Commons Attribution-ShareAlike/GNU Free Documentation License terms, and YAGO3 data under Creative Commons Attribution terms. Our derived datasets, code, and checkpoints are provided in the anonymous repository for research reproducibility and evaluation. We do not claim a separate license for these derived artifacts in this submission, and users should comply with the original licenses and access conditions of the underlying KGs when using or redistributing them. We use these resources only for entity alignment research, benchmark construction, and model evaluation, which is consistent with their intended use as public knowledge resources.

We do not collect or process private data. However, public KGs may contain entities corresponding to real people, organizations, places, and creative works, and may therefore include names or descriptions of identifiable public entities. We do not infer private attributes, add new personal information, or link KG entities to non-public records. Our processing is limited to public KG identifiers, labels, and 1-hop relational neighborhoods already present in the source dumps. Since the datasets are intended for EA research, they should not be used to make consequential decisions about individuals or to construct private identity-resolution systems without additional review.

A main ethical risk in EA is false entity merging: incorrect alignments can propagate errors in downstream KG fusion and may merge distinct real-world entities that share the same name, such as different people named ``Jim Jones.'' Our hard-negative benchmarks are designed to surface this risk, and HELEA aims to reduce it by grounding decisions in relational structure rather than name similarity. However, the HELEA-Reranker relies on an LLM and may still produce biased, uncertain, or hallucinated judgments. Therefore, in downstream applications, its outputs should be treated as candidates for human review rather than ground truth. Finally, applying same-name hard-negative mining to private KGs may reveal sensitive disambiguation information, so users should verify compliance with relevant data governance policies before deployment.

\bibliography{custom}

\appendix

\section{Implementation Details}
\label{app:impl}

The entity encoder (all-MiniLM-L6-v2~\cite{minilm, reimers-gurevych-2019-sentence}) is fine-tuned for 43,000 steps with batch size 512, learning rate $2 \times 10^{-5}$, 4,300 warmup steps, and maximum input length of 256 tokens.
For DW, 10,000 samples are held out from \textbf{DW-Train} as a validation set; the entity encoder trains on the remaining 5.56M samples.
For DY, a separate entity encoder is trained on \textbf{DY-Train} (2.0M samples) under identical hyperparameters with 10,000 samples held out.
These held-out sets are used exclusively for the $\alpha$ ablation (Section~\ref{sec:ablation_alpha}) and threshold selection. No test benchmark data is used in any tuning decision.
Entity representations consist of the entity name prepended to its deduplicated 1-hop triple sequence.
We set $K{=}10$ based on HELEA-Retriever's Hit@10 of 0.995 on DW-15K (Table~\ref{tab:retrieval_full}), indicating the true positive is present in the top-10 candidates for 99.5\% of queries; increasing $K$ further adds LLM inference cost while recovering fewer than 0.5\% of remaining cases.
Top-10 candidates are reranked by Gemma 4 31B Instruct~\cite{gemma4_2025} served via vLLM~\cite{kwon2023efficient} in standard instruction-following mode (thinking disabled).

\section{Full Retrieval Results on DW-15K and DY-15K}
\label{app:retrieval_full}

Table~\ref{tab:retrieval_full} reports the complete retrieval metrics (Hit@1, Hit@5, Hit@10, MRR) for all methods on DW-15K and DY-15K.
The main paper (Table~\ref{tab:exp1}) reports only Hit@1 and MRR. This table provides the full breakdown for reference.

\begin{table*}[!t]
\centering
\renewcommand{\arraystretch}{1.2}
\resizebox{0.9\textwidth}{!}{%
\begin{tabular}{l|cccc|cccc}
\hline
& \multicolumn{4}{c|}{\textbf{DW-15K}} & \multicolumn{4}{c}{\textbf{DY-15K}} \\
\textbf{Method} & \textbf{Hit@1} & \textbf{Hit@5} & \textbf{Hit@10} & \textbf{MRR} & \textbf{Hit@1} & \textbf{Hit@5} & \textbf{Hit@10} & \textbf{MRR} \\ \hline
\multicolumn{9}{l}{\textit{Name-dependent baselines}} \\ \hline
String-matching      & 0.825 & 0.825 & 0.825 & 0.826 & 0.999 & 0.999 & 0.999 & 0.999 \\
Name-only Bi-encoder & 0.899 & 0.939 & 0.944 & 0.918 & 0.993 & 0.999 & 0.999 & 0.996 \\ \hline
\multicolumn{9}{l}{\textit{Reproduced baselines}} \\ \hline
BERT-INT~\cite{Bertint}        & \textbf{0.994} & \textbf{0.995} & \textbf{0.995} & \textbf{0.994} & \textbf{0.999} & \textbf{0.999} & \textbf{0.999} & \textbf{0.999} \\
SelfKG~\cite{SelfKG}           & 0.920 & 0.947 & 0.954 & 0.931 & \underline{0.998} & \underline{0.999} & \underline{0.999} & \underline{0.999} \\
ChatEA~\cite{ChatEA}           & 0.955 & 0.965 & 0.966 & 0.961 & 0.980 & 0.986 & 0.986 & 0.983 \\
UniEA~\cite{UniEA}             & 0.901 & 0.939 & 0.947 & 0.919 & 0.760 & 0.880 & 0.899 & 0.814 \\
HLMEA~\cite{hlmea_2025}        & 0.842 & 0.917 & 0.931 & 0.877 & 0.989 & 0.997 & 0.998 & 0.993 \\ \hline
\multicolumn{9}{l}{\textit{Ours}} \\ \hline
HELEA-Retriever & 0.987 & 0.994 & 0.995 & 0.991 & 0.992 & 0.998 & 0.999 & 0.995 \\
\textbf{HELEA}  & \underline{0.993} & \underline{0.995} & \underline{0.995} & \underline{0.994} & 0.992 & 0.998 & 0.999 & 0.995 \\ \hline
\end{tabular}%
}
\caption{Full retrieval results on DW-15K and DY-15K (Hit@1, Hit@5, Hit@10, MRR).
  Evaluation protocol follows Table~\ref{tab:exp1}.
  \textbf{Bold}/\underline{underline}: best/second-best per column.}
\label{tab:retrieval_full}
\end{table*}

\section{Dataset Statistics}
\label{app:stats}

Table~\ref{tab:name_overlap} reports the exact name-overlap counts for each benchmark split,
and Table~\ref{tab:benchmark_stats} gives complete size statistics for all training corpora
and evaluation sets.

Table~\ref{tab:name_overlap} confirms that our hard-negative benchmarks eliminate the name-bias shortcut: in DW-HN29K, 99.07\% of hard negatives share an exact name with their positive counterpart, rising to 100.00\% in DY-HN27K.
The latter reflects YAGO's complete derivation of entity labels from Wikipedia titles---identical to those used by DBpedia---making every DY-HN27K decision a purely structural disambiguation task.

Table~\ref{tab:benchmark_stats} reveals an asymmetry between the two KG pairs. DW-Train (5.57M samples) is nearly three times the size of DY-Train (1.99M), primarily because Wikidata contains far more within-KG same-name entity pairs than YAGO (2.22M Wikidata:Wikidata vs.\ 455K YAGO:YAGO hard negatives).
The positive-to-negative ratio also differs: DW-Train is slightly negative-heavy (2.48M / 3.09M), whereas DY-Train is positive-heavy (1.45M / 0.54M), reflecting the different degrees of name collision in each KG.

\begin{table*}[!t]
\centering
\renewcommand{\arraystretch}{1.2}
\resizebox{0.9\textwidth}{!}{%
\begin{tabular}{l|cccc}
\hline
& \textbf{DW-15K} & \textbf{DW-HN29K} & \textbf{DY-15K} & \textbf{DY-HN27K} \\ \hline
Positive pairs & 15,000 & 14,718 & 15,000 & 12,635 \\
\quad w/ exact name match & 12,257 (81.71\%) & 12,065 (81.97\%) & 14,988 (99.92\%) & 12,635 (100.00\%) \\
Negative pairs & random & 15,000 (hard) & random & 15,000 (hard) \\
\quad w/ exact name match & n/a & 14,861 (99.07\%) & n/a & 15,000 (100.00\%) \\
\hline
\end{tabular}%
}
\caption{Name overlap statistics in DW-15K, DW-HN29K, DY-15K, and DY-HN27K.}
\label{tab:name_overlap}
\end{table*}

\section{Notes on DY Dataset Characteristics}
\label{app:dy_results}

Two structural features of the DY dataset family make it qualitatively different from DW.

\paragraph{DY-15K is effectively a name-matching test.}
With 99.92\% name overlap in positive pairs (Table~\ref{tab:name_overlap}), DY-15K leaves almost no room for structure-aware evaluation: String-matching achieves Hit@1 0.999 and Name-only Bi-encoder achieves 0.993.
HELEA-Retriever reaches Hit@1 0.992, and adding the HELEA-Reranker yields no further change---when the retriever already surfaces the name-identical true positive as its top candidate in virtually all queries, the reranker's listwise comparison produces identical rankings.
This is expected behavior on a near-ceiling name-matching task; the reranker's contribution becomes apparent on DY-HN27K, where names are entirely uninformative and the +21.2\%p gain (F1 0.719 $\to$ 0.931) is driven solely by triple-based disambiguation.
DY-15K retrieval scores should therefore not be interpreted as a measure of relational understanding.
Note also that YAGO stores human-readable labels in triple tails, making entity names implicitly visible.

\paragraph{YAGO triple sparsity limits structural representation quality.}
As discussed in Section~\ref{sec:exp2}, YAGO's average of 8.79 triples per entity yields sparser context, explaining why HELEA's DY-15K Hit@1 (0.992) trails BERT-INT (0.999).

\paragraph{DY-HN27K is the more informative benchmark.}
Since 100\% of hard negatives share an exact name with their positive, DY-HN27K forces every decision to rely solely on relational structure.
Despite sparse triples, the HELEA-Reranker substantially recovers: HELEA achieves F1 0.931, a gain of +21.2~pp over HELEA-Retriever (F1 0.719).
Among reproduced baselines, ChatEA achieves the strongest competing result (F1 0.920) through full-context LLM reasoning, confirming that LLM-based disambiguation is key to handling same-name entities even in sparse-triple settings (see Section~\ref{sec:exp3}).

\begin{table*}[!t]
\centering
\renewcommand{\arraystretch}{1.2}
\resizebox{0.7\textwidth}{!}{%
\begin{tabular}{l|rr}
\hline
\textbf{Split} & \textbf{Samples} & \textbf{Positive / Negative} \\ \hline
DW-HN29K (benchmark)      & 29,718    & 14,718 / 15,000 \\
DW-15K   (retrieval eval) & 15,000    & 15,000 / 0 \\
DY-HN27K (benchmark)      & 27,635    & 12,635 / 15,000 \\
DY-15K   (retrieval eval) & 15,000    & 15,000 / 0 \\ \hline
DW-Train (training)     & 5,568,150 & 2,480,293 / 3,087,857 \\
DY-Train (training)     & 1,996,008 & 1,452,392 / 543,616 \\ \hline
Hard negatives (DBpedia:Wikidata)  & 2,701     & --- \\
Hard negatives (DBpedia:DBpedia)   & 1,378,235 & --- \\
Hard negatives (Wikidata:Wikidata) & 2,220,511 & --- \\
Hard negatives (DBpedia:YAGO)      & 88,566    & --- \\
Hard negatives (YAGO:YAGO)         & 455,050   & --- \\ \hline
\end{tabular}%
}
\caption{Dataset statistics for all benchmarks and training corpora.}
\label{tab:benchmark_stats}
\end{table*}
\section{GPT-oss-120B Score Fusion Ablation}
\label{app:gpt_oss}

To assess whether the $\alpha$-sweep findings from Section~\ref{sec:ablation_alpha} generalize
beyond the primary Gemma~4~31B Instruct backbone, we repeat the same experiment with
GPT-oss-120B as the HELEA-Reranker backbone (Table~\ref{tab:gpt_oss}).
HELEA-Retriever and all other settings are kept identical. Only the reranker backbone changes.
The trends mirror those observed with Gemma~4~31B Instruct (Table~\ref{tab:exp5}).
On DW-HN29K, the retriever collapses at $\alpha = 1.0$ (F1 0.854) and improves steadily
to F1 0.940 with HELEA-Reranker-only scoring (+8.6\%p).
On DY-HN27K, the retriever achieves F1 0.719 at $\alpha = 1.0$; GPT-oss-120B reranking
yields a substantial gain to F1 0.892 at $\alpha = 0.00$ (+17.3\%p), though this trails
Gemma~4~31B Instruct's gain of +21.2\%p, suggesting a relative gap in GPT-oss-120B's
instruction-following calibration on the listwise ranking format for the DY setting.
Absolute scores are lower than with Gemma~4~31B Instruct (DW F1 0.940 vs.\ 0.967;
DY F1 0.892 vs.\ 0.931), but the qualitative pattern---any HELEA-Reranker presence substantially
recovers collapse caused by name bias---holds consistently across backbones.
The residual gap likely reflects differences in instruction-following capability for the listwise ranking format, particularly in assigning calibrated confidence scores across 10 candidates simultaneously.
Taken together, these results confirm that HELEA is model-agnostic: any sufficiently capable LLM can serve as the HELEA-Reranker and substantially recover from collapse caused by name bias, without retraining the entity encoder.

\begin{table}[!t]
\centering
\renewcommand{\arraystretch}{1.2}
\resizebox{0.95\columnwidth}{!}{%
\begin{tabular}{l|cc|cc}
\hline
& \multicolumn{2}{c|}{\textbf{DW-HN29K}} & \multicolumn{2}{c}{\textbf{DY-HN27K}} \\
\textbf{$\alpha$} & \textbf{Acc.} & \textbf{F1}
                  & \textbf{Acc.} & \textbf{F1} \\ \hline
1.00 (Retriever) & 0.844 & 0.854 & 0.724 & 0.719 \\
0.75             & 0.902 & 0.895 & 0.901 & 0.889 \\
0.50             & 0.939 & 0.935 & 0.900 & 0.891 \\
\textbf{0.25}             & \textbf{0.942} & \textbf{0.940} & \textbf{0.903} & 0.891 \\
0.00 (Reranker) & 0.942 & 0.940 & 0.901 & \textbf{0.892} \\ \hline
\end{tabular}%
}
\caption{Score fusion ablation with GPT-oss-120B as the HELEA-Reranker backbone,
         evaluated on the DW-HN29K and DY-HN27K test sets.
         $\alpha{=}1.0$: HELEA-Retriever only; $\alpha{=}0.0$: HELEA-Reranker (GPT-oss-120B) only.
         \textbf{Bold} marks the best value per column.}
\label{tab:gpt_oss}
\end{table}

\section{LLM Reranking Prompt Template}
\label{app:prompt}

Figure~\ref{fig:prompt_template} shows the full prompt template used for \textit{listwise} reranking across all evaluation tasks.
The model receives the query entity alongside the top-10 retrieved candidates, each with their names, serialized 1-hop KG context, retriever rank, and cosine score, and outputs a ranked list of \texttt{index:score} pairs.
For the classification benchmarks (DW-HN29K, DY-HN27K), the same listwise prompt is used; the confidence score assigned to the target evaluation pair is then extracted from the output and fused with the entity encoder score for binary prediction.

\begin{figure*}[!t]
\centering
\small
\begin{tcolorbox}[colback=gray!5!white, colframe=gray!60!black,
    title=\textbf{LLM Reranking Prompt Template},
    width=\textwidth, boxrule=0.8pt]
\begin{verbatim}
[System]
Base your judgment on the entity names and knowledge graph triples.
Entities with identical names can refer to completely different real-world objects
— use the relational structure to disambiguate.

[Query]
Entity A
birthPlace: United States | occupation: politician | party: Democratic Party | ...

[Candidate 1]  (Retriever Rank: 1, Score: 0.924)
birthPlace: United States | occupation: politician | party: Republican Party | ...

[Candidate 2]  (Retriever Rank: 2, Score: 0.911)
genre: jazz | occupation: musician | instrument: piano | ...

...

[Candidate 10]  (Retriever Rank: 10, Score: 0.841)
country: France | sport: football | position: forward | ...

Rank the candidates from best to worst match for Entity A.
You MUST begin your response immediately with 'RANKING:'.
Respond strictly in the following format:

RANKING: i:<score>, j:<score>, k:<score>, ...
(scores are confidence values in [0, 1]; e.g., RANKING: 3:0.92, 1:0.75, 2:0.41, ...)
Reasoning: <2-3 sentences citing the discriminating triples>
\end{verbatim}
\end{tcolorbox}
\caption{Prompt template used for \textit{listwise} LLM reranking across all evaluation benchmarks.
For classification benchmarks, the confidence score for the target evaluation pair is extracted from this same listwise output and thresholded for binary prediction.
Actual entity triples are substituted in place of the placeholders shown above.}
\label{fig:prompt_template}
\end{figure*}

\end{document}